\RequirePackage{amsmath}
\documentclass[runningheads]{llncs}
\usepackage{makeidx}  % allows for indexgeneration
\usepackage{graphicx}
\usepackage{cite}
\usepackage{cleveref}
\usepackage[bottom]{footmisc}

\titlerunning{Real-Time Detection of Early Squamous Neoplasia}  % abbreviated title (for running head)
\title{Interpretable Fully Convolutional Classification of Intrapapillary Capillary Loops for Real-Time Detection of Early Squamous Neoplasia}

\author{
	Luis C. Garcia-Peraza-Herrera \inst{1}
	Martin Everson \inst{2}
	Wenqi Li \inst{1}
	Inmanol Luengo \inst{1}
	Lorenz Berger \inst{5}
	Omer Ahmad \inst{1} 
	Laurence Lovat \inst{2}
	Hsiu-Po Wang \inst{3} 
	Wen-Lun Wang \inst{4}
	Rehan Haidry \inst{2}
	Danail Stoyanov \inst{1}
	Tom Vercauteren \inst{1}
	Sebastien Ourselin \inst{1}
}

\authorrunning{Luis C. Garc\'ia-Peraza-Herrera et al.} % abbreviated author list (for running head)
\institute{
Wellcome / EPSRC Centre for Interventional and Surgical Sciences, London, UK\\
\and
University College London Hospitals, London, UK\\
\and
Gastroenterology National Taiwan University, Taipei, Taiwan\\
\and
E-Da Cancer Hospital, Kaohsiung, Taiwan\\
\and
Innersight Labs, London, UK\\
}

\begin{document}
\maketitle              % typeset the title of the contribution
\begin{abstract}
In this work, we have concentrated our efforts on the interpretability of classification results coming from a fully convolutional neural network. Motivated by the classification of oesophageal tissue for real-time detection of early squamous neoplasia, the most frequent kind of oesophageal cancer in Asia, we present a new dataset and a novel deep learning method that by means of deep supervision and a newly introduced concept, the embedded Class Activation Map (eCAM), focuses on the interpretability of results as a design constraint of a convolutional network. We present a new approach to visualise attention that aims to give some insights on those areas of the oesophageal tissue that lead a network to conclude that the images belong to a particular class and compare them with those visual features employed by clinicians to produce a clinical diagnosis. In comparison to a baseline method which does not feature deep supervision but provides attention by grafting Class Activation Maps, we improve the F1-score from $87.3$\% to $92.7$\% and provide more detailed attention maps. 

\end{abstract}

\section{Introduction}

\begin{figure}[t!]
	\centering
	\includegraphics[width=\textwidth]{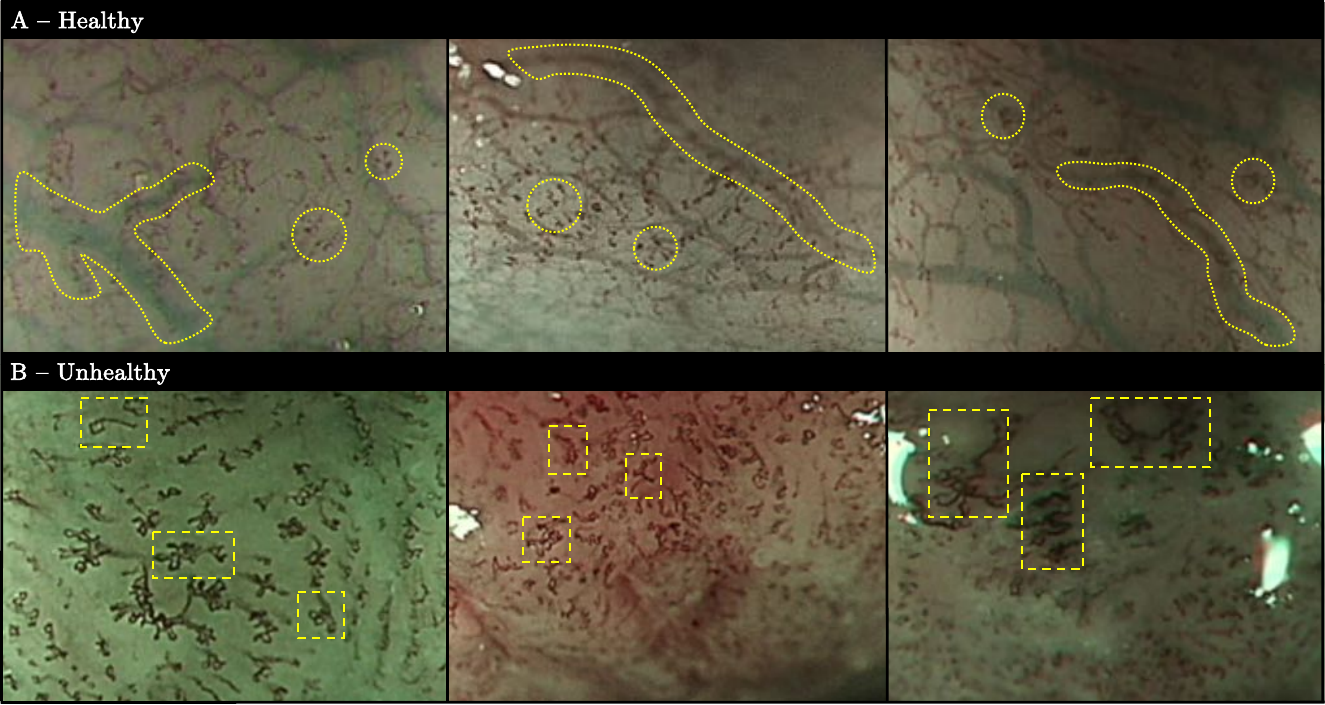}
	\caption{Narrow-Band Imaging magnifying endoscopy of the oesophagus. (A) Frames extracted from surveillance endoscopies on several healthy subjects. In these patients mucosal vessels can be easily observed (dotted). In addition, interpapillary capillary loops (circles) are perceived as minuscule dots connected to an extremely thin filament. (B) Images from patients with abnormal interpapillary capillary loops suggesting carcinoma depth invasion. Microvessels are dilated and present unusual shape irregularities (rectangles).}
	\label{fig:ipcl}
\end{figure}

Motivated by the clinical problem of intrapapillary capillary loops (IPCL) classification, we introduce a novel dataset containing 7046 frames from 17 patients (see table \ref{tab:dataset} for more details), and a novel unified framework for automatic feature extraction, classification and visual interpretability of results. We present a novel convolutional network architecture that focuses on the interpretability of the results as a design constraint for the network and serves as a baseline for quantitative comparison of results with future methods. We compare the visual features highlighted in the heatmaps produced by the network with those that are clinically relevant to produce a diagnosis.

In the Computer-Assisted Interventions (CAI) community labelled data is often scarce. 
%Obtaining it is very costly for several reasons that are challenging to circumvent, long waiting times to obtain ethical approval, difficulties to recruit patients and production of annotations being incredibly time consuming. 
Deep learning has become extremely popular due to its success in tasks such as classification and segmentation, but a large amount of data is typically required to capture the variability of the data across patients.
%to produce models that are able to capture the variability of a particular problem across patients.
As researchers in this area, we are also faced with additional challenges. Clinical collaborators are interested in interpreting the results coming from computer-assisted systems. That is, understanding the process followed by deep learning approaches to make a diagnosis. This includes analysing which features present in the images lead to a certain output and if those coincide with the ones that they analyse during clinical examination of the data. Conversely, it is also interesting to discover whether automatically extracted features are different from the ones currently used in clinical practice but can nonetheless lead to a correct diagnosis.

%Interpretability of the data plays also a key role in more practical aspects. 
%In addition to scarcity of labelled data, there are several other reasons why one could end up working with small datasets. First, clinical collaborators tend to expect results in small pilot studies before going for large scale endeavours. Second, it is a common practice to create small datasets such as mini-ImageNet \cite{Vinyals2016a}, as developing deep learning methods based on large datasets is an incredibly time-consuming task. While is true that technical solutions to this problem exist, as training ImageNet in one hour with 256 GPUs \cite{Vinyals2016a}, these resources are rarely available in academic environments. 

Using reduced datasets can potentially lead to models that do not generalise well. While it is true that there are efforts to build large scale CAI databases \cite{Maier-Hein2017a}, in this paper, we have concentrated our efforts on the interpretability of classification results coming from a fully convolutional neural network trained on a small dataset. 
% As a pretext task, we introduce a new problem, the classification of oesophageal tissue for detection of early squamous carcinoma. 

%Most of the superficial oesophagus lining is formed by squamous cells, which when become malignant  
Squamous cell carcinoma (SCC) is the most frequent kind of oesophageal cancer in Asia \cite{Wang2004}, presenting rapidly increasing numbers in the western world in recent decades too. Early diagnosis -and resection- play a key role to increase the chances of survival \cite{Endo1997}, as superficial lesions present low rates of lymphatic dissemination. Detection is currently achieved by screening programs on high risk populations \cite{Wang2004}. Narrow-Band Imaging magnifying endoscopy (NBI-ME) is the state-of-the-art technique employed for screening \cite{Oyama2017}. In addition to early diagnosis, a precise estimation of depth of invasion is crucial. Lesions that are closer to the oesophageal surface (mucosal layer) can be treated by minimally invasive endoscopic therapy rather than surgery \cite{Ono2006}.

NBI-ME facilitates the visualisation of micro-vascular patterns, called intrapapillary capillary loops (IPCL), which are linked to early squamous cell carcinoma and present focal, subtle, and easily missed visual features, particularly in centres with a low amount of cases. It has been also shown that the thickness and tortuosity of IPCL patterns is highly correlated with histological state and depth of invasion \cite{Oyama2017}. Hence, having an automated red-flag system that analyses each video frame in real-time could potentially help detect subtle IPCL patterns that might be difficult to distinguish by unspecialised endoscopists (see figure \ref{fig:ipcl}).

Recent work has explored different approaches to analyse the implicit attention mechanisms of convolutional neural networks. In \cite{Given2013}, authors produce attention heatmaps as a linear combination of feature maps from the last convolutional layer. The weighting coefficients are extracted from the fully connected output neurons. Zhou et al. \cite{Zhou2015} allow for a fully convolutional classification by means of Global Average Pooling (GAP). When GAP is omitted (at inference), instead of a vector of class probabilities, a Class Activation Map (CAM) is automatically generated. In addition, these maps enable for accurate object localization, a task for which the network has not been trained for. %More recently, Guan et al. \cite{Guan2018} generate attention heatmaps by means of a \textit{maxout} operation across the feature maps produced by the last convolutional layer.

\section{Materials and Methods}

\subsection{Deeply Supervised Embedded Class Activation Maps (eCAM)}

There are various reasons why a fully convolutional classification is convenient for the proposed pretext task. Different endoscope processors provide images of varying resolutions. We aim for a flexible method that can generalise to different input sizes to simplify the preprocessing of data. It is also of interest to give the method versatility to process both full images or cropped patches. Furthermore, there are images of the oesophagus that can present unhealthy IPCL patterns only in certain areas. Hence, we seek for a method that could potentially have the ability to classify seamlessly both images and patches. GAP \cite{Zhou2015} has been shown as a feasible way to reduce the feature maps to a single value and still maintain a state-of-the-art classification accuracy.

As interpretability of the results is of utmost relevance, in addition to the classification score, we aim to obtain an attention heatmap that exposes to workings of the inference process and highlights those visual features that led the network conclude an image belongs to a certain class. This is relevant because it helps to check whether the network is paying attention to those parts of the image that clinical experts consider to be determinant to produce a valid diagnosis. Furthermore, it serves as a validation mechanism that could point out possible problems in the learning process, for example, in case the network pays attention to areas of the image that are clinically irrelevant but happen to contain discriminative spurious visual features. 

\begin{figure}[b!]
	\centering
	\includegraphics[width=\textwidth]{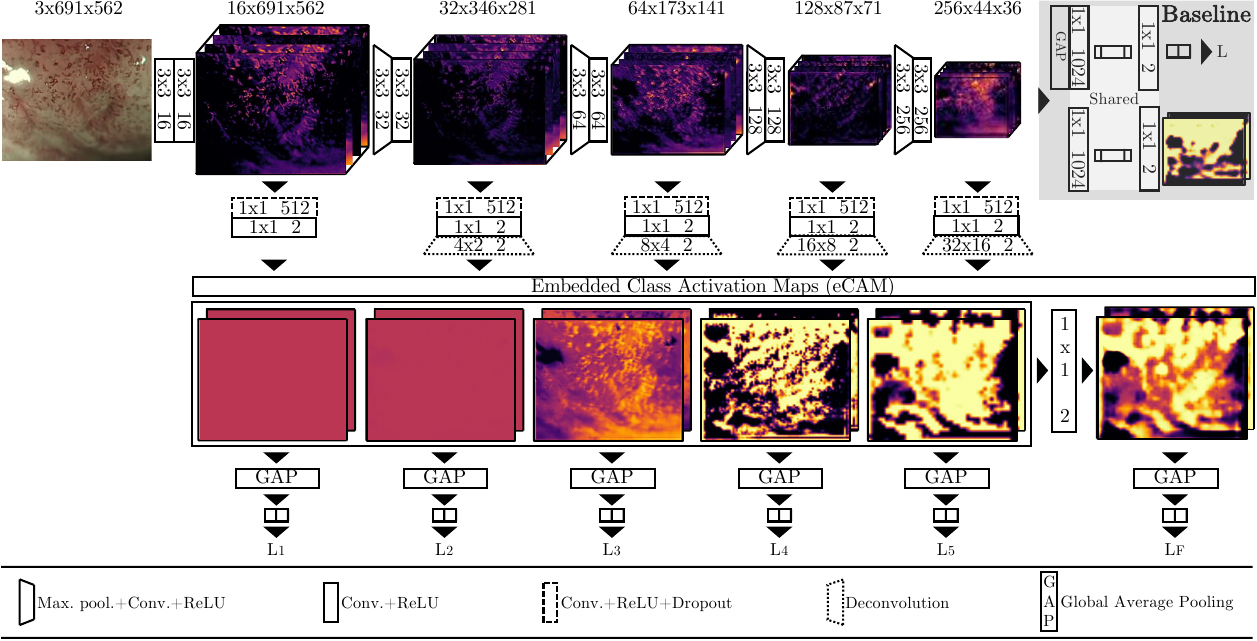}
	\caption{Proposed convolutional network with multi-scale embedded Class Activation Maps. Baseline architecture (i.e. without deep supervision and classical CAM) shown shaded.}
	\label{fig:network}
\end{figure}

Class Activation Maps \cite{Zhou2015} are a recent attempt to produce meaningful attention heatmaps. They have shown to produce comparable to state-of-the-art localizations without re-training for the task. However, with this approach and the baseline network presented in figure \ref{fig:network}, we achieve low resolution attention heatmaps that might allow to find a large object in scenes of daily life but lack the definition to illustrate attention in oesophageal NBI-ME images.
In \cite{Garcia-Peraza-Herrera2017}, the authors show that deep supervision is able to achieve superior results on segmentation of medical images. The reasons to opt for a similar approach are three-fold: fast convergence as gradients flow quicker to early layers of the network, ability to produce predictions at different resolutions, and improved quantitative classification results. Furthermore, we can take advantage of deep supervision to produce high resolution attention heatmaps (based on learnt deconvolutions that upsample the heatmaps to the original image size). As opposed to \cite{Zhou2015}, we do not only aim to generate heatmaps that show the implicit attention of the network, but want to explicitly force the architecture to produce one attention map per class and use them to generate a classification prediction. We therefore introduce a new architecture that is fully convolutional, produces embedded attention heatmaps that allow for clinical interpretation of the results and works in real-time. The proposed method is shown in figure \ref{fig:network}. The deep supervision mechanism is composed by several losses. As can be observed in figure \ref{fig:network}, a loss $L_s(\hat{\theta})$ is calculated using cross-entropy at each resolution scale:
\begin{align}
L_s(\hat{\boldsymbol{\theta}}) = - \sum_{k=1}^{K} g^{nk} \log{\hat{p}_s^{nk}}
\end{align}
where $\hat{\boldsymbol{\theta}}$ are the network weights, $s \in \{1, ..., 5\}$, $g^{nk}$ is one when observation $n$ belongs to class $k$ and zero otherwise, and $\hat{p}_s^{nk}$ is the predicted probability for observation $n$ belonging to class $k$ at scale $s$. $L_F(\hat{\boldsymbol{\theta}})$ (see figure \ref{fig:network}) has the same structure as $L_s(\hat{\boldsymbol{\theta}})$ but the probability for each class comes from applying GAP to a weighted sum of the eCAM at all scales. That is, all the scale-dependent attention heatmaps are fused by means of a learnt 1x1 convolution so that the attention information at all scales is employed to produce a classification prediction. The final training loss to be minimised during the training process is
\begin{align}
\mathcal{L}(\hat{\boldsymbol{\theta}}) = \frac{1}{S + 1}\biggl(L_F(\hat{\boldsymbol{\theta}}) + \sum_{s=1}^{S}L_s(\hat{\boldsymbol{\theta}})\biggr)
\end{align}
where $S = 5$ scales.

\subsection{Dataset}

As we are introducing a new clinical problem, no dataset exists for the IPCL classification task. This novel dataset originates from 17 monocular videos (one video per patient) captured with two NBI-ME systems, \textsc{Olympus Lucera CV-260 \& CV-290} (Olympus Corporation, Tokyo, Japan). These videos have been recorded during routine screenings, and depict oesophageal recordings, starting on the stomach pit and finishing on the upper oesophageal sphincter. The oesophagus is cleaned prior to examination to expose clearly the mucosa.

The video sequences are cut to extract the useful parts of the procedure and sampled at 30fps as the endoscopists tend to perform rapid movements with the camera. The extracted frames are then quality controlled by an expert to discard those images that do not allow to perform a diagnosis with confidence. The final images are cropped so that no black corners or borders are left. All the pixels belong to oesophageal tissue. The labels have been matched with histological results from biopsies performed during the screening. The dataset contains 7046 frames, whose resolution ranges from 458x308 to 696x308 pixels. The dataset has been divided in three subsets, training, validation and testing. Each subset contains frames from different patients. To perform a thorough evaluation, five cross-validation folds have been created. Each fold contains a different draw of patients (see table \ref{tab:dataset} for more details). 

All procedures performed in studies involving human participants were in accordance with the ethical standards of the local institutions and with the 1964 Helsinki declaration and its later amendments or comparable ethical standards. %The dataset is available at Anonymized.

\begin{table}[t!]
	\centering
	\caption{Number of frames for each cross-validation split. For each fold, the frames in the training, validation and testing sets belong to different patients. The letters (A) and (B) indicate whether the frames belong to class A -healthy- or B -unhealthy-.} 
	\begin{tabular}{lcccccc}
		\hline
		\multicolumn{1}{c}{\bfseries ~Fold No.~} & \multicolumn{1}{c}{\bfseries ~Train. (A)~} & \multicolumn{1}{c}{\bfseries ~Val. (A)~} & \multicolumn{1}{c}{\bfseries ~Test. (A)~} & \multicolumn{1}{c}{\bfseries ~Train. (B)~} & \multicolumn{1}{c}{\bfseries ~Val. (B)~} & \multicolumn{1}{c}{\bfseries ~Val. (B)~} \\ \hline	
		
		1 &  2620  &  201  &  577  &  2803  &  258   &  587 \\	
		
		2 &  1792  &  891  &  715  &  2205  &  739   &  704 \\
		
		3 &  1822  &  685  &  891  &  1549  &  1360  &  739 \\
		
		4 &  1792  &  715  &  891  &  1912  &  961   &  775 \\
		
		5 &  1559  &  685  &  1154  &  1754  &  743  &  1151 \\ \hline
		
		\textbf{Average}  &  1917  &  635  &  646  &  2045  &  812 &  791 \\ \hline
		
	\end{tabular}
	\vspace{0.2cm}
	\label{tab:dataset}
\end{table}

\subsection{Implementation details}
Stochastic Gradient Descent (SGD) was the optimizer of choice. The training was performed with a fixed learning rate across training of $1e-6$, momentum of $0.9$, and a batch size of $1$. A different CNN is trained for each dataset fold. All the networks are trained with a maximum number of iterations of $4\times$ the number of images in the fold's training set. Training weights are saved every $200$ iterations and the best performing snapshot in validation set is selected for testing. \textsc{caffe} 1.0.0-rc5 \cite{Jia2014} with \textsc{CUDNN} 5.1.10, \textsc{CUDA} 8.0.61, and \textsc{NVIDIA} driver 384.111 was the deep learning setup for development. The experiments were run on an Intel Core i7-4790K CPU @ 4.00GHz and an \textsc{NVIDIA} GeForce TITAN X (Pascal).

\section{Results and Discussion}

The proposed method achieves an average sensitivity and F1-score across dataset folds of $89.7$\% and $92.7$\% respectively in comparison with the baseline sensitivity and F1-score of $82.7$\% and $87.3$\% (see detailed quantitative evaluation on table \ref{tab:results}). As shown in \cite{Xie2015} deep supervision boosts accuracy by forcing the network to learn discriminant features at different resolutions. This is particularly relevant for endoscopy as features are visible or not depending on the distance from the camera to the oesophageal wall and the network has to be able to learn not only which features are useful at each scale but also how to fuse the predictions from different resolutions to achieve a correct classification. The prediction time interval ranges from $26.17$ms for the smallest images in the dataset to $37.48$ms for the largest ones.

\begin{table}[t!]
	\centering
	\caption{Testing set classification results for the unhealthy class. Sensitivity, specificity, accuracy, precision and F1-score are reported.} 
	\begin{tabular}{l|ccccc|ccccc}
		\hline
		& \multicolumn{5}{c|}{\bfseries Baseline} & \multicolumn{5}{c}{\bfseries Proposed} \\
		\hline
		\multicolumn{1}{c|}{\bfseries ~Fold No.~} & \multicolumn{1}{c}{\bfseries ~Sens.~} & \multicolumn{1}{c}{\bfseries ~Spec.~} & \multicolumn{1}{c}{\bfseries ~Acc.~} & \multicolumn{1}{c}{\bfseries ~Prec.~} & \multicolumn{1}{c|}{\bfseries ~F1~} & \multicolumn{1}{c}{\bfseries ~Sens.~} & \multicolumn{1}{c}{\bfseries ~Spec.~} & \multicolumn{1}{c}{\bfseries ~Acc.~} & \multicolumn{1}{c}{\bfseries ~Prec.~} & \multicolumn{1}{c}{\bfseries ~F1~}\\ \hline
		
		1 &  77.5 & 99.8 & 88.6 & 99.8 & 87.2 & 80.4   &  92.0  &  86.2  &  91.1  &  85.4 \\
		
		2 &  39.2 & 99.9 & 69.8 & 99.6 & 56.3 & 78.1   &  99.7  &  89.0  &  99.6  &  87.6 \\
		
		3 &  100.0 & 97.1 & 98.4 & 96.6 & 98.3 & 100.0  &  95.9  &  97.7  &  95.2  &  97.6 \\
		
		4 &  96.6 & 97.9 & 97.3 & 97.5 & 97.1 & 99.4   &  97.3  &  98.3  &  97.0  &  98.2 \\
		
		5 &  100.0 & 95.6 & 97.8 & 95.8 & 97.8 & 90.6   &  99.6  &  95.1  &  99.5  &  94.9 \\ \hline
		
		\textbf{Average} & 82.7 & 98.0 & 90.4 & 97.9 & 87.3 & 89.7  &  96.9  &  93.3  &  96.5  &  92.7 \\ \hline
		
	\end{tabular}
	\vspace{0.2cm}
	\label{tab:results}
\end{table}

\begin{figure}[t!]
	\centering
	\includegraphics[width=.97\textwidth]{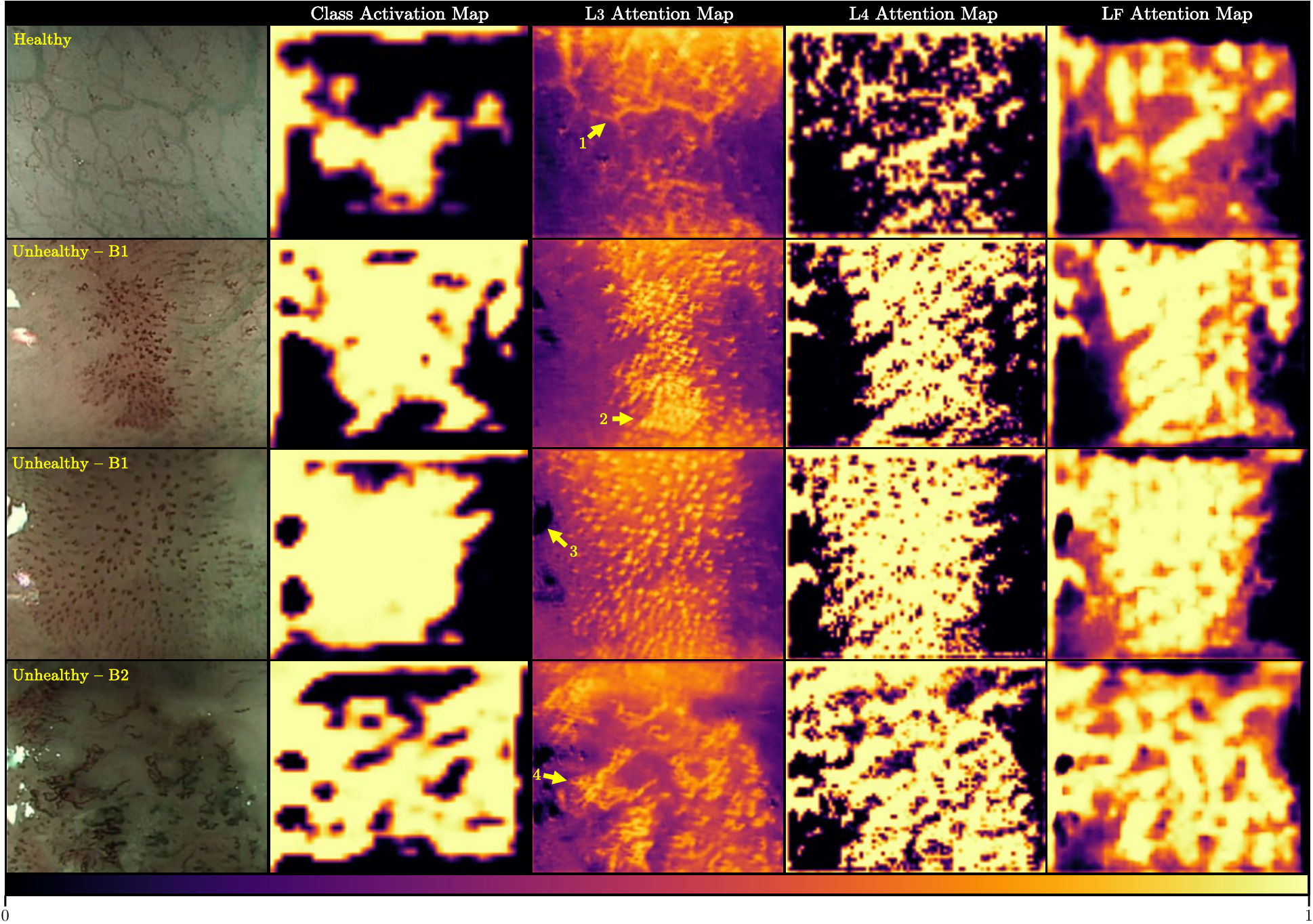}
	\caption{Deeply supervised embedded Class Activation Maps. (1) Mucosal vessel erroneously highlighted in $L_3$. (2) Densely populated IPCL area highlighted. (3) Specular reflection discarded. (4) \textit{Star-shaped} irregular IPCL pattern highlighted.}
	\label{fig:results}
\end{figure}

In figure \ref{fig:results} we show different video frames and their corresponding eCAM. Only those from resolution levels L3, L4 and L5 and the multi-scale fused version are shown. The eCAM of resolution scale L1 and L2 are highly uncertain, as can be observed in figure \ref{fig:network}. The reason possibly being two-fold. First, it is too early in the network and there are not enough filters (design constraint to achieve real-time) to capture the complexity of the disease. Second, receptive field being too small to capture discriminative visual features at those resolutions. 

Figure \ref{fig:results} shows several interesting visual features captured by eCAM. Specular reflections which are uninformative for tissue classification are discarded. The heatmaps are able to highlight both global and disperse unhealthy IPCL patterns and focalized areas of diseased tissue.  Furthermore, despite recognising vessels as a matter of attention, the network does not seem to be able to discern that thick mucosal vessels are benign in healthy frames. Irregular \textit{star-shaped} severe IPCL patterns (shown in the last row of figure \ref{fig:results}) are also successfully highlighted as diseased. Large areas of healthy tissue are successfully recognised as benign as can be observed in the first row of figure \ref{fig:results}.

%High resolution deeply supervised eCAM seem as a promising technique to provide high resolution attention maps that highlight relevant visual features despite training a CNN for tissue classification. Despite forcing the network for which there is no label classification accuracy is not affected.
%
%The combination of deeply supervised eCAM at multiple resolutions along with GAP as a way to generate class scores from attention heatmaps achieve improved quantitative results compared to a classifier trained with just a traditional encoder and GAP as mean to achieve both attention analysis (through CAM) and fully convolutional classification.

\section{Conclusion}

Motivated by the problem of oesophageal IPCL pattern binary classification (healthy vs. unhealthy), we presented the first publicly available dataset for the task. We proposed a novel deeply supervised convolutional architecture that performs real-time fully convolutional classification achieving an average F1-score of 92.7\%. We introduced the concept of embedded Class Activation Maps (eCAM) as a technique to force the network to capture and store visual attention maps and use them as source of information for classification. We showed that by means of deep supervision it is possible to obtain high quality heatmaps at the original resolution of the image. Future work will focus on the extension to multi-class detection of different unhealthy IPCL patterns.

%\subsubsection{Acknowledgements.}
%This work was supported by Wellcome Trust [WT101957], EPSRC (NS/A000027/1, EP/H046410/1, EP/J020990/1, EP/K005278), NIHR BRC UCLH/UCL High Impact Initiative and a UCL EPSRC CDT Scholarship Award (EP/L016478/1). The authors would like to thank NVIDIA for the donated GeForce GTX TITAN X GPU, their colleagues E. Maneas, S. Moriconi, F. Chadebecq, M. Ebner and S. Nousias for the ground truth of \texttt{FetalFlexTool} and E. Maneas for preparing setup with an \textit{ex vivo} placenta.

\bibliographystyle{splncs}
\bibliography{library}

\end{document}